\documentclass[final,5p,times,twocolumn]{elsarticle}

\usepackage{amsmath,amsfonts}
\usepackage{algorithmic}
\usepackage{array}
\usepackage{stfloats}
\usepackage{verbatim}
\usepackage{graphicx}
\usepackage{booktabs} 
\usepackage{multirow} 
\usepackage{qcircuit}
\usepackage{amsmath}
\usepackage{graphicx}
\usepackage{braket}
\usepackage{float}

\usepackage{mathtools}
\DeclarePairedDelimiter\ceil{\lceil}{\rceil}

\usepackage{amssymb}
\usepackage{amsthm}
\usepackage{lineno}
\usepackage[ruled,linesnumbered,lined]{algorithm2e}
\usepackage{amsmath}
\usepackage[utf8]{inputenc}
\usepackage{longtable}
\usepackage{caption}
\usepackage{subcaption}
\usepackage{morefloats}
\usepackage{float}

\begin{document}

\begin{frontmatter}
\title{Data Complexity Measures for Quantum Circuits Architecture Recommendation}

\author[ufpe]{Fernando M. de Paula Neto}
        \address[ufpe]{Centro de Inform\'{a}tica \\ Universidade Federal de Pernambuco}



\begin{abstract}
Quantum Parametric Circuits are constructed as an alternative to reduce the size of quantum circuits, meaning to decrease the number of quantum gates and, consequently, the depth of these circuits. However, determining the optimal circuit for a given problem remains an open question. Testing various combinations is challenging due to the infinite possibilities. In this work, a quantum circuit recommendation architecture for classification problems is proposed using database complexity measures. 
A quantum circuit is defined based on a circuit layer and the number of times this layer is iterated. Fourteen databases of varying dimensions and different numbers of classes were used to evaluate six quantum circuits, each with 1, 2, 3, 4, 8, and 16-layer repetitions. Using data complexity measures from the databases, it was possible to identify the optimal circuit capable of solving all problems with up to 100$\%$ accuracy. Furthermore, with a mean absolute error of 0.80 $\pm$ 2.17, one determined the appropriate number of layer repetitions, allowing for an error margin of up to three additional layers. Sixteen distinct machine learning models were employed for the selection of quantum circuits, alongside twelve classical regressor models to dynamically define the number of layers.
\end{abstract}


\end{frontmatter}

\section{Introduction}
Quantum computing (QC) leverages principles from quantum mechanics to perform information processing. In addition to exploring intrinsically quantum phenomena such as superposition and entanglement~\cite{simon1997power}, QC becomes even more relevant as the miniaturization of electronic components reaches the atomic level, and the laws of quantum mechanics come into play to operate them. The different ways of operating its calculations have allowed advantages to be explored about existing classical algorithms. There are quadratic gains in relation to known classical algorithms, as seen in Grover's search algorithm applied to disordered arrays \cite{grover1996fast}, as well as exponential gains, exemplified by Shor's algorithm for prime number factorization \cite{shor}. Quantum supremacy, wherein it is possible to demonstrate that quantum computers can solve tasks more effectively than known classical algorithms, has sparked significant interest and exploration in the field of quantum computing \cite{arute2019quantum}. Experimentally, other quantum achievements have gained merit over their classical counterparts \cite{bravyi2020quantum}. The field of machine learning also has potential to benefit from quantum attributes \cite{riste2017demonstration, biamonte2017quantum}.

Quantum processors currently operate in the era of noisy intermediate-scale quantum (NISQ). This implies that processors with only a few quantum bits are available, accompanied by pronounced noise. Such limitations suggest that theoretical quantum algorithms, due to their numerous quantum gates (i.e., deep circuits), may not be practically deployable \cite{liu2020reliability}. In light of this, a way to circumvent the creation of small quantum circuits is by utilizing a quantum circuit framework in which quantum gates are parameterized, delegating the optimization of these parameters for a given task to an external algorithm. In this framework, parameterized quantum circuits (PQCs) are created to perform classification, predictions, and approximations, while a classical machine is used to update the circuit parameters \cite{preskill2018quantum, mcclean2016theory}. For this reason, often the system as a whole is referred to as a \textit{hybrid quantum-classical framework}.


Several options are available for tackling the development and/or design of PQCs, given the infinite range of possibilities. Hence, a persistent challenge persists.
Options include associating the computational power of circuits with their accuracy capacity \cite{coyleborn, schuld2008effect}; as well as exploring correlations between circuit descriptors (such as entanglement level and circuit expressivity) to guide the choice and design of PQCs \cite{sim2019expressibility, hubregtsen2021evaluation}. Constructive methods that use iterative processes are also employed in the creation of such circuits \cite{tang2021qubit}.

In this work, it is proposed that measures extracted from the database \cite{lorena2019complex} be used as a recommendation  for the choice of a quantum circuit, including its quantum circuit layout and the repetition frequency of this layout. To the best of the author's knowledge, there is no prior research on the choice of database information for the selection of quantum circuits.
This type of strategy that takes into account information extracted from data is considered a meta-learning approach, and it has proven to be useful in choosing classification models in various contexts \cite{ostlund2022data, okimoto2017complexity}.

This article is organized as follows: In Section \ref{sec:complexmeasures}, database complexity measures are presented. In Section \ref{sec:qc}, the basic principles and mathematical notation of quantum computing are presented. The experimental protocol is presented in Section \ref{sec:exp}. The experimental results and their discussion are presented in Section \ref{sec:results}. Conclusions and future work are presented in Section \ref{sec:conclusion}.

\section{Complexity database metrics}
\label{sec:complexmeasures}

In this section, 22 measures of database complexity are presented, along with an explanation of the information extracted from the data. The measures can be categorized into six groups. \textbf{Feature-based measures} assess the informativeness of available features in distinguishing between classes. \textbf{Linearity measures} aim to quantify the potential for linear separation among classes. \textbf{Neighborhood measures} provide insights into the presence and density of either the same or different classes in local neighborhoods. \textbf{Network measures} extract structural information from the dataset by modeling it as a graph. \textbf{Dimensionality measures} evaluate data sparsity by considering the number of samples relative to the data dimensionality. \textbf{Class imbalance measures} account for the ratio of the number of examples between classes.

\subsection{Feature-based measures}
\subsubsection{Maximum Fisher’s Discriminant Ratio (F1)}
The maximum Fisher’s discriminant ratio \cite{ho2002complexity,sotoca2006meta}, denoted here by F1, measures the overlap between the values of the features in different classes.
Low values in the F1 measure suggest the presence of at least one feature with minimal overlap among different classes. This indicates the existence of a feature for which a hyperplane perpendicular to its axis can effectively separate the classes. 

\subsubsection{The Directional-vector Maximum Fisher’s Discriminant Ratio (F1v)}
This measure is used 
as a complement to F1 \cite{orriols2010dcol}. It searches for a vector that can separate the two classes after the examples have been projected into it and considers a directional Fisher criterion. Smaller F1v values, constrained within the (0, 1] interval, signify simpler classification problems. In such instances, a linear hyperplane can effectively segregate a significant portion, if not all, of the data, aligning appropriately with the feature axes. 

\subsubsection{Volume of Overlapping Region (F2)}
The F2 measure \cite{lorena2012analysis, ho2002complexity} assesses the degree of overlap in the distributions of feature values within classes. A higher F2 value indicates increased overlap among classes, reflecting higher complexity in the problem. Conversely, an F2 value of zero is expected when there is at least one non-overlapping feature.

\subsubsection{Maximum Individual Feature Efficiency (F3)}

This measure estimates the individual efficiency of each feature in separating the classes, and considers the maximum value found among the features \cite{ho2002complexity}. 
Here, one considers the complement of this measure to assign higher values to more complex problems. For each feature, the assessment involves checking for value overlap among examples from different classes. If overlap exists, the classes are deemed ambiguous in that region. 

\subsubsection{Collective Feature Efficiency (F4)}

The F4 measure
offers insight into how features collaborate. It follows a sequential process akin to F3, beginning with the selection of the most discriminative feature — i.e., the feature exhibiting minimal overlap between different classes \cite{orriols2010dcol}. 
Lower F4 values signify the potential discrimination of more examples, indicating a simpler problem. 

\subsection{Linearity Measures}
\subsubsection{Sum of the Error Distance by Linear Programming (L1)}
This measure evaluates the linear separability of data by calculating, for a given dataset, the sum of distances from incorrectly classified examples to a linear boundary used in their classification \cite{ho2002complexity}. If the L1 value is zero, it indicates that the problem is linearly separable, suggesting a simpler scenario compared to problems requiring a non-linear boundary.

\subsubsection{Error Rate of Linear Classifier (L2)}

The L2 measure computes the error rate of the linear SVM classifier \cite{ho2002complexity}.
Higher L2 values denote more errors and therefore a greater complexity regarding the aspect that the data cannot be separated linearly. 

\subsubsection{Non-Linearity of a Linear Classifier (L3)}

The procedure begins by generating a new dataset through the interpolation of pairs of training examples from the same class \cite{ho2002complexity}. In this process, two randomly chosen examples from the same class undergo linear interpolation, yielding a new example. Subsequently, a linear classifier is trained on the original data, and its error rate is assessed using the newly generated data points. This index is sensitive to the distribution of class data in border regions and the extent of overlap between the convex hulls that delineate the classes. Notably, it identifies the presence of concavities in class boundaries.
Higher values in this index signify increased complexity.

\subsection{Neighborhood measures}
\subsubsection{Fraction of Borderline Points (N1)}
In this metric, a Minimum Spanning Tree (MST) is initially constructed from the data. Each vertex corresponds to an example, and the edges are weighted based on the distance between them. N1 is determined by calculating the percentage of vertices incident to edges connecting examples from opposing classes in the generated MST \cite{ho2002complexity}.
N1 serves as an estimate for the size and complexity of the required decision boundary, identifying critical points in the dataset—those nearby but belonging to different classes. Higher N1 values indicate a greater demand for complex boundaries to separate classes and/or a substantial degree of overlap between the classes.

\subsubsection{Ratio of Intra/Extra Class Nearest Neighbor Distance (N2)}
This metric calculates the ratio of two sums: (i) the sum of distances between each example and its closest neighbor from the same class (intra-class); and (ii) the sum of distances between each example and its closest neighbor from a different class (extra-class) \cite{ho2002complexity}.

Low N2 values signify simpler problems, where the overall distance between examples of different classes exceeds the overall distance between examples from the same class. N2 exhibits sensitivity to the distribution of data within classes, considering not only the characteristics of the boundary between classes but also the overall internal structure. 

\subsubsection{Error Rate of the Nearest Neighbor Classifier (N3)}

The N3 measure refers to the error rate of a 1NN classifier that is estimated using a leave-one-out procedure \cite{ho2002complexity}. 
High N3 values indicate that many examples are close to examples of other classes, making the problem more complex. 

\subsubsection{Non-Linearity of the Nearest Neighbor Classifier (N4)}
This metric bears similarity to L3 but employs the Nearest Neighbor (NN) classifier instead of the linear predictor \cite{ho2002complexity}. Elevated N4 values signal more complex problems. Unlike L3, N4 can be directly applied to multiclass classification problems, eliminating the necessity to decompose them into binary subproblems initially.

\subsubsection{Fraction of Hyperspheres Covering Data (T1)}

The concept involves obtaining a maximum-order adherence subset for each example, comprising only instances from the same class \cite{ho2002complexity}. Subsets entirely encompassed by other subsets are discarded. 
The measure captures not only the distribution near the class boundary but also the overall distribution of data within the classes.

\subsubsection{Local Set Average Cardinality (LSC)}
The Local-Set (LS) of an example xi within a dataset (T) comprises points from T whose distance to xi is less than the distance from xi to its nearest enemy \cite{leyva2014set}. The cardinality of an example's LS signifies its proximity to the decision boundary and the narrowness of the gap between classes. Consequently, examples separated from the other class with a narrow margin will exhibit a lower LS cardinality.

\subsection{Network measures}

\subsubsection{Average density of the network (Density)}

This metric assesses the normalized number of retained edges in a graph constructed from the dataset, divided by the maximum number of edges between n pairs of data points \cite{garcia2015effect}. Lower values in this metric indicate dense graphs where many examples are interconnected. Such a scenario is typical for datasets with dense regions from the same class, suggesting lower complexity. 

\subsubsection{Clustering coefficient (ClsCoef)}

The clustering coefficient measure assesses the grouping tendency of the graph vertexes, by monitoring how close to form cliques neighborhood vertexes are \cite{garcia2015effect}. It will be smaller for simpler datasets, which will tend to have dense connections among examples from the same class.

\subsubsection{Hub score (Hubs)}
The hub score assigns a score to each node based on its number of connections to other nodes, weighted by the number of connections these neighbors possess \cite{garcia2015effect}. Consequently, highly connected vertices linked to other highly connected vertices will garner a higher hub score, serving as an indicator of each node's influence in the graph.
In complex datasets where significant class overlap is present, robust vertices may tend to be less connected to similarly robust neighbors. Conversely, in simpler datasets characterized by dense regions within classes, higher hub scores are anticipated.

\subsection{Class balance measures}
\subsubsection{Average number of features per points (T2)}
T2 divides the dataset's number of examples by their dimensionality \cite{ho2002complexity}.
T2 serves as an indicator of data sparsity, capturing scenarios where numerous predictive attributes are present, but few data points result in a sparse distribution within the input space. Lower T2 values signify reduced sparsity, indicating simpler problems due to the absence of low-density regions that might impede the induction of an effective classification model.

\subsubsection{Average number of PCA dimensions per points (T3)}

The metric T3 employs Principal Component Analysis (PCA) on the dataset \cite{ho2002complexity}. Unlike T2, which utilizes the raw dimensionality of the feature vector, T3 relies on the number of PCA components required to capture 95\% of the data variability as the foundation for assessing data sparsity. Similar to T2, smaller T3 values are indicative of simpler datasets with lower sparsity.

\subsection{Ratio of the PCA Dimension to the Original Dimension (T4)}
This metric provides an approximate gauge of the proportion of relevant dimensions within the dataset \cite{ho2002complexity}. Relevance is assessed using the PCA criterion, aiming for a transformation of features into uncorrelated linear functions that effectively capture most of the data variability. A higher T4 value indicates a greater necessity for the original features to describe the data variability.

\subsection{Entropy of class proportions (C1)}
The C1 metric serves to quantify the imbalance within a dataset. It attains its minimum value for balanced problems, where all class proportions are equal \cite{ho2002complexity}. Such problems are deemed simpler based on the aspect of class balance

\subsection{Imbalance ratio (C2)}
The C2 measure is a well-known index computed for measuring class balance \cite{tanwani2008classification}. 
Larger values of C2 are obtained for imbalanced problems. The minimum value of C2 is achieved for balanced problems.

\begin{table*}[H]
\begin{tabular}{lrrr}
\hline
\textbf{Category}                        & \textbf{Name}                                          & \textbf{Acronym} & \textbf{Ref.} \\ 
\midrule
\multirow{5}{*}{Feature overlapping}     & Maximum Fisher’s Discriminant Ratio                    & F1               &               \\ 
                                         & Directional-vector maximum Fisher’s discriminant ratio & F1v              &               \\  
                                         & Volume of the overlapping region                       & F2               &               \\ 
                                         & Maximum individual feature efficiency                  & F3               &               \\ 
                                         & Collective feature efficiency                          & F4               &               \\ \hline
\multirow{3}{*}{Linearity measures}      & Sum of the error distances by linear programming       & L1               &               \\ 
                                         & Error rate of a linear classifier                      & L2               &               \\ 
                                         & Non-linearity of a linear classifier                   & L3               &               \\ \hline
\multirow{6}{*}{Neighborhood measures}   & Fraction of borderline points                          & N1               &               \\  
                                         & Ratio of intra/inter class nearest neighbor distance   & N2               &               \\ 
                                         & Error rate of the nearest neighbor classifier          & N3               &               \\
                                         & Non-linearity of the nearest neighbor classifier       & N4               &               \\  
                                         & Fraction of hyperspheres covering data                 & T1               &               \\  
                                         & Local set average cardinality                          & LSCAvg           &               \\ \hline
\multirow{3}{*}{Network measures}        & Average density of the network                         & Density          &               \\  
                                         & Clustering coefficient                                 & ClsCoef          &               \\  
                                         & Hub score                                              & Hubs             &               \\ \hline
\multirow{3}{*}{Dimensionality measures} & Average number of points per dimension                 & T2               &               \\  
                                         & Average number of points per PCA dimension             & T3               &               \\ 
                                         & Ratio of the PCA dimension to the raw dimension        & T4               &               \\ \hline
\multirow{2}{*}{Class balance measures}  & Entropy of class proportions                           & C1               &               \\ 
                                         & Imbalance ratio                                        & C2               &               \\ \bottomrule
\end{tabular}
\end{table*}

\section{Quantum computing}
\label{sec:qc}

 \subsection{Quantum bits}
    The unit of information in the quantum computation is called a \emph{quantum bit} (qubit). The qubit is a two-dimensional vector in the complex vector space $\mathbb{C}^2$. It can be in superposition of states, \textit{ie.} in position $0$ or position $1$ at the same time, if one considers the canonical basis as $0$ and $1$. Any qubit $\ket{\psi}$ can be written as a linear combination of vectors (or states) of $\mathbb{C}^2$ canonical (or computational) basis $\ket{0} = [1,0]^T$ and $\ket{1} = [0,1]^T$  as viewed in Equation ~\ref{eq:qubit}, 
    \begin{equation}
    \label{eq:qubit}
    \ket{\psi} = \alpha\ket{0} + \beta\ket{1}
    \end{equation}
    where $\alpha$ and $\beta$ are complex number and $|\alpha|^2+|\beta|^2=1$. This notation also means that the qubit has $|\alpha|^2$ to be measured as $0$ and $|\beta|^2$ to be measured as $1$. Throughout the text of this article, the symbol $i$ will denote the complex number $\sqrt{-1}$.
    
    The qubits are represented mathematically united by the tensor operator, $\otimes$. The tensor operator is used to represent quantum systems with two or more qubits $\ket{\textbf{g}} = \ket{ab} = \ket{a} \otimes \ket{b}$. Here one will use the bold font for the representation of quantum states with more than one qubit.
    For two qubits $\ket{a} = \alpha_1 \ket{0} + \beta_1 \ket{1}$ and $\ket{b} = \alpha_2 \ket{0} + \beta_2 \ket{1}$, the tensor operator generates the state $\ket{ab} = \alpha_1 \alpha_2 \ket{00} + \alpha_1 \beta_2 \ket{01} + \beta_1 \alpha_2 \ket{10} + \beta_1 \beta_2 \ket{11}$. For a general two states $\ket{\textbf{p}}$ and $\ket{\textbf{q}}$ with $n$ and $m$ states respectively, the state $\ket{pq}$ can be calculated by the operation described in Equation \ref{eq:TensorProduct}.
    \begin{equation}
        \left[
    \begin{array}{ll}
    \alpha_1 \\
    \alpha_2 \\
    ... \\
    \alpha_{2^n}
    \end{array}
    \right]
    \otimes
        \left[
    \begin{array}{ll}
    \beta_1 \\
    \beta_2 \\
    ... \\
    \beta_{2^m}
    \end{array}
    \right]
    = \left[
    \begin{array}{ll}
    \alpha_1 \left[
    \begin{array}{ll}
    \beta_1 \\
    \beta_2 \\
    ...\\
    \beta_{2^m}
    \end{array}
    \right]     \\ \\
    \alpha_2 \left[
    \begin{array}{ll}
    \beta_1 \\
    \beta_2 \\
    ...\\
    \beta_{2^m}
    \end{array}
    \right] \\ 
    ...\\
    \alpha_{2^n} \left[
    \begin{array}{ll}
    \beta_1 \\
    \beta_2 \\
    ...\\
    \beta_{2^m}
    \end{array}
    \right]
    \end{array}
    \right]     
    =\left[
    \begin{array}{ll}
    \alpha_1 \beta_1 \\
    \alpha_1 \beta_2 \\
    ...\\
    \alpha_1 \beta_{2^m}\\
    \alpha_2 \beta_1\\
    \alpha_2 \beta_2\\
    ...\\
    \alpha_2 \beta_{2^m}\\
    ... \\
    \alpha_{2^n} \beta_{2^m}
    \end{array}
    \right]
    \label{eq:TensorProduct}
    \end{equation}
    
    One can represent the quantum states using integer numbers rather than string bits inside the $\ket{.}$ notation. For a given quantum state with $n$ states $\ket{\boldsymbol{\psi}} = \alpha_1 \ket{1} + \alpha_2 \ket{2} + \cdot + \alpha_{n} \ket{n}$ the measurement of the $\ket{x}$ state may occur with $|\bra{\boldsymbol{\psi}} \ket{x}|^2$ of probability where the $\bra{.}$ represents the complex conjugate of the vector $\ket{.}$.
    
    Let $Q$ and $R$ be two vector spaces the tensor product of $Q$ and $R$, denoted by   $Q\otimes R$, is the vector space generated by the tensor product of all vectors $\ket{a}\otimes \ket{b}$, with $\ket{a} \in A$ and $\ket{b} \in B$. Some states $\ket{\psi}\in Q \otimes R$  cannot be written as a product of states of its component systems $Q$ and $R$. States with this property are called \emph{entangled} states.
    %
    
    \subsection{Quantum operators}
    
    The quantum states are modified by quantum operators which change the amplitude values of the qubits. A quantum operator $\mathbf{U}$ 
    over 
    $n$ qubits is a unitary complex matrix 
    of order 
    $2^n\times 2^n$. For example, some operators over
    1 qubit are Identity $\mathbf{I}$, NOT $\mathbf{X}$, and Hadamard $\mathbf{H}$, described below in Equation~\eqref{eq:quantumop1} and Equation~\eqref{eq:quantumop2} in matrix form and operator form. The combination of these unitary operators forms a quantum circuit.
    \begin{equation}
    \label{eq:quantumop1}
    \begin{array}{ll}
    \textbf{I}= \left[
    \begin{array}{ll}
    1 & 0\\
    0 & 1\\
    \end{array}
    \right]
    \begin{array}{l}
    \textbf{I}\ket{0}= \ket{0} \\
    \textbf{I}\ket{1}=\ket{1} 
    \end{array}
    \end{array}
    \begin{array}{ll}
    \textbf{X}= \left[
    \begin{array}{ll}
    0 & 1\\
    1 & 0\\
    \end{array}
    \right]
    \begin{array}{l}
    \textbf{X}\ket{0}= \ket{1} \\
    \textbf{X}\ket{1}=\ket{0} 
    \end{array}
    \end{array}
    \end{equation}
    \begin{equation}
    \label{eq:quantumop2}
    \begin{array}{ll}
    \textbf{H}=\frac{1}{\sqrt{2}} \left[
    \begin{array}{cc}
    1 & 1\\
    1 & -1\\
    \end{array}
    \right]
    &
    \begin{array}{l}
    \textbf{H}\ket{0}= 1/\sqrt{2}(\ket{0}+\ket{1}) \\
    \textbf{H}\ket{1}=1/\sqrt{2}(\ket{0}-\ket{1}) 
    \end{array}
    \end{array}
    \end{equation}
    The Identity operator $\mathbf{I}$ generates the output exactly as the input; $\mathbf{X}$ operator works as the classic NOT in the computational basis; Hadamard $\mathbf{H}$ generates a superposition of states when applied on a computational basis.

    In the same way one can combine quantum states, quantum operators can also be combined using tensor product. For two $(n_0, m_0)$-dimensional matrix $U$ and $(n_1, m_1)$-dimensional matrix $V$, their composition, $U \otimes V$, products a third $(n_0 n_1,m_0 m_1)$-dimensional matrix. One denotes as $\mathbf{A}^{\otimes {s}}$ the $s$-fold application of $\mathbf{A}$.

Any single-qubit quantum gate can be described as a combination of gates $\mathbf{Rx}$, $\mathbf{Ry}$, and $\mathbf{Rz}$, described respectively in Equations \ref{eq:Rx}, \ref{eq:Ry}, and \ref{eq:Rz}, as follows $U~=~\mathbf{Rx}(\alpha) \mathbf{Ry}(\beta) \mathbf{Rz}(\gamma)$, or even in the form described by Equation \ref{eq:U}.

\begin{equation}
  \mathbf{Rx}(\theta) = \begin{pmatrix}
    cos(\theta/2) & -i \cdot sin(\theta/2)\\ 
    -i \cdot sin(\theta/2) & cos(\theta/2)
  \end{pmatrix}
  \label{eq:Rx}
\end{equation}

\begin{equation}
  \mathbf{Ry}(\theta) = \begin{pmatrix}
    cos(\theta/2) & -sin(\theta/2)\\ 
    sin(\theta/2) & cos(\theta/2)
  \end{pmatrix}
  \label{eq:Ry}
\end{equation}

\begin{equation}
  \mathbf{Rz}(\theta) = \begin{pmatrix}
    e^{-i \cdot \theta/2} & 0\\ 
    0 & e^{i \cdot \theta/2}
  \end{pmatrix}
  \label{eq:Rz}
\end{equation}

\begin{equation}
  \mathbf{U}(\theta, \beta, \gamma) = \begin{pmatrix}
    cos(\theta/2) & -e^{i \cdot \gamma} \cdot sin(\theta)\\ 
    e^{i \cdot \beta} sin(\theta) & e^{i \cdot (\beta + \gamma)} cos(\theta)
  \end{pmatrix}
  \label{eq:U}
\end{equation}

The $\mathbf{CNOT}$ is a two qubits operator. It has a control qubit and a target qubit. It works considering the value of the control qubit to apply the $\mathbf{X}$ operator on the target qubit. If  the control qubit is set to $1$ the $\mathbf{X}$ operator is applied to the target qubit. The matrix representation for the computational basis  is shown in Equation \ref{eq:CNOTworking}. 

\begin{equation}
\label{eq:CNOTworking}
\begin{array}{ll}
\mathbf{CNOT} = 
\begin{bmatrix}
 1 & 0 & 0 & 0\\
 0 & 1 & 0 & 0\\
 0 & 0 & 0 & 1\\
 0 & 0 & 1 & 0\\
\end{bmatrix}
&
\begin{array}{l}
\mathbf{CNOT}\ket{00}= \ket{00} \\
\mathbf{CNOT}\ket{01}= \ket{01} \\
\mathbf{CNOT}\ket{10}= \ket{11} \\
\mathbf{CNOT}\ket{11}= \ket{10} 
\end{array}
\end{array}
\end{equation}

It can be generalized and defined as an $(n+1)-$ary $\mathbf{CNOT}$ having $n$ control qubits and requiring all the control qubits to be $\ket{1}$ for applying $\mathbf{X}$. Another even more significant generalization is that any gate can be controlled, just like the $\mathbf{CNOT}$, i.e., a C-$\mathbf{U}$ gate. The operation is similar to the $\mathbf{CNOT}$ gate, where the $\mathbf{U}$ gate is applied to the target qubit if the control qubit is in the value 1. With all the single-qubit quantum gates and the CNOT it is possible to build any quantum algorithm \cite{nielsen2010quantum}.

\subsection{Quantum measurement}

\begin{figure}[h]
\[
\Qcircuit @C=.7em @R=.3em {
\lstick{\ket{0}} & \qw & \gate{H} & \qw & \meter 
}
\]
\caption{Measurement results in $0$ or $1$ with equal probability $\frac{1}{2}$.}
\label{fig:meas}
\end{figure}
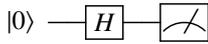

Measurement is an irreversible operation that partially or totally loses the information about the superposition of states. For a qubit state $\ket{\psi} = \alpha \ket{0} + \beta \ket{1}$, a measurement collapses (projects) the state either to $\ket{0}$ state with $|\alpha|^2$ of probability or to $\ket{0}$ with $|\beta|^2$ probability. For a composed and superposed quantum state $\ket{\textbf{g}}$ a probability to see a state $\ket{\textbf{i}}$ is $|{\langle g | i \rangle}|^2$. In Figure \ref{fig:meas}, an example of the measurement operator in a quantum circuit is shown.

The mean expected value $\braket{Z}$ of this measurement for an arbitrary qubit $\ket{\psi}$ is $\braket{Z} = \bra{\psi}Z\ket{\psi}$, where $Z = \begin{bmatrix}
 1 & 0 \\
 0 & -1 \\ 
\end{bmatrix}$. This value is between -1 and 1. For more details on quantum measurements, see \cite{nielsen2010quantum}. 

\subsection{Quantum circuit}

One can represent quantum operations by quantum circuits. This graphical representation considers the qubits as wires and quantum operators as boxes. The flow of the execution, as in the classical case, is from left to right.

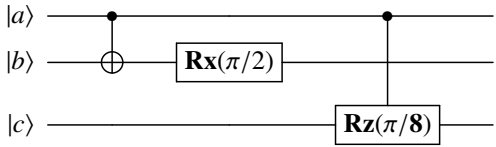
\begin{figure}[H]
\[
\Qcircuit @C=2em @R=0.9em {
\lstick{\ket{a}} & \ctrl{1} & \qw & \ctrl{2} & \qw\\
\lstick{\ket{b}} & \targ & \gate{\mathbf{Rx}(\pi/2)} & \qw & \qw \\
\lstick{\ket{c}} & \qw & \qw & \gate{\mathbf{Rz(\pi/8)}} & \qw
}
\]
\caption{An example of quantum circuit with one $\mathbf{CNOT}$ operator and two rotation gates, $\mathbf{Rx}$ and $\mathbf{Rz}$, one of them being controlled (controlled rotation).}
\label{fig:cnot}
\end{figure}

Figure~\ref{fig:cnot} has an example of a quantum circuit composed of a $\mathbf{CNOT}$,  where the control qubit is depicted by a filled circle and a $\mathbf{X}$ operator. There is also a controlled $\mathbf{Rz}$ gate, as well as an $\mathbf{Rx}$ gate.

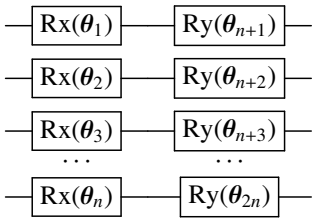
\begin{figure}[H]
    \centering
\[ \Qcircuit @C=1em @R=.5em {
& \gate{\text{Rx($\boldsymbol{\theta}_{1}$)}} & \qw & \gate{\text{Ry($\boldsymbol{\theta}_{n+1}$)}} & \qw \\
& \gate{\text{Rx($\boldsymbol{\theta}_{2}$)}} & \qw & \gate{\text{Ry($\boldsymbol{\theta}_{n+2}$)}} & \qw \\
& \gate{\text{Rx($\boldsymbol{\theta}_{3}$)}} & \qw & \gate{\text{Ry($\boldsymbol{\theta}_{n+3}$)}} & \qw \\
&  \cdots                                       &  &                                                  \cdots \\ 
& \gate{\text{Rx($\boldsymbol{\theta}_{n}$)}} & \qw & \gate{\text{Ry($\boldsymbol{\theta}_{2n}$)}} &\qw 
} \]
    \caption{Layer C-1.}
    \label{fig:layerC1}
\end{figure}

\begin{figure}[H]
    \centering
\[ \Qcircuit @C=1em @R=.5em {
& \gate{\text{Ry($\boldsymbol{\theta}_{1}$)}}   & \qw & \ctrl{1} & \qw & \qw      & \qw      & \qw \\
& \gate{\text{Ry($\boldsymbol{\theta}_{2}$)}}   & \qw & \targ    & \qw & \ctrl{1} & \qw      & \qw \\
& \gate{\text{Ry($\boldsymbol{\theta}_{3}$)}}   & \qw & \qw      & \qw &  \targ   & \qw      & \qw \\
&  \cdots                                         &  &                                                  \cdots \\ 
& \gate{\text{Ry($\boldsymbol{\theta}_{n-1}$)}} & \qw & \qw      & \qw & \qw      & \ctrl{1} & \qw\\
& \gate{\text{Ry($\boldsymbol{\theta}_{n}$)}}   & \qw & \qw      & \qw & \qw      & \targ    & \qw
} \]
    \caption{Layer C-2.}
    \label{fig:layerC2}
\end{figure}
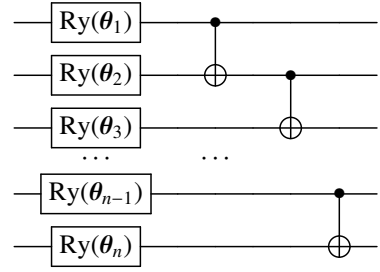

\begin{figure}[H]
    \centering
\[ \Qcircuit @C=0.17em @R=.5em {
& \gate{\text{Ry($\boldsymbol{\theta}_{1}$)}}   & \qw & \ctrl{1} & \qw & \qw      & \qw      & \qw \\
& \gate{\text{Ry($\boldsymbol{\theta}_{2}$)}}   & \qw &  \gate{\text{Rz($\boldsymbol{\theta}_{n+1}$)}}    & \qw & \ctrl{1} & \qw      & \qw \\
& \gate{\text{Ry($\boldsymbol{\theta}_{3}$)}}   & \qw & \qw      & \qw &  \gate{\text{Rz($\boldsymbol{\theta}_{n+2}$)}}   & \qw      & \qw \\
&  \cdots                                         &  &                                                  \cdots \\ 
& \gate{\text{Ry($\boldsymbol{\theta}_{n-1}$)}} & \qw & \qw      & \qw & \qw      & \ctrl{1} & \qw\\
& \gate{\text{Ry($\boldsymbol{\theta}_{n}$)}}   & \qw & \qw      & \qw & \qw      & \gate{\text{Rz($\boldsymbol{\theta}_{2n-1}$)}}    & \qw
} \]
    \caption{Layer C-3.}
    \label{fig:layerC3}
\end{figure}
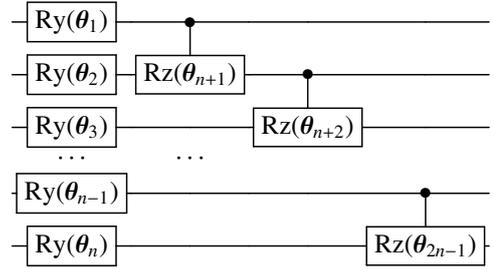

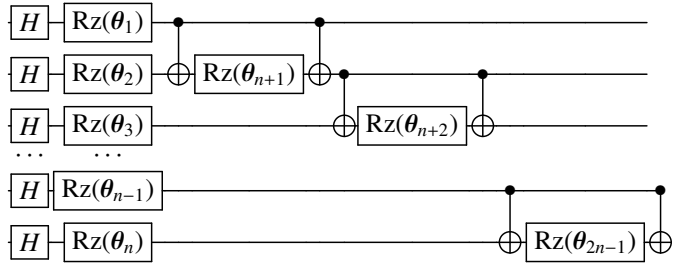
\begin{figure*}
    \centering
\[ \Qcircuit @C=0.1em @R=.5em {
& \gate{H} & \qw & \gate{\text{Rz($\boldsymbol{\theta}_{1}$)}}   & \qw & \ctrl{1} & \qw & \qw                                          & \qw & \ctrl{1}        &  \qw                                          & \qw & \qw      & \qw  & \qw & \qw  & \qw      & \qw & \qw \\
& \gate{H} & \qw & \gate{\text{Rz($\boldsymbol{\theta}_{2}$)}}   & \qw &  \targ   & \qw & \gate{\text{Rz($\boldsymbol{\theta}_{n+1}$)}} & \qw &\targ & \ctrl{1} &  \qw                                          & \qw & \ctrl{1} & \qw  & \qw & \qw  & \qw      & \qw \\
& \gate{H} & \qw & \gate{\text{Rz($\boldsymbol{\theta}_{3}$)}}   & \qw & \qw      & \qw &  \qw                                         & \qw & \qw  & \targ    &  \gate{\text{Rz($\boldsymbol{\theta}_{n+2}$)}} & \qw & \targ    &  \qw & \qw & \qw  & \qw      & \qw \\
&  \cdots                                         &  &                                                  \cdots \\ 
& \gate{H} & \qw & \gate{\text{Rz($\boldsymbol{\theta}_{n-1}$)}} & \qw & \qw      & \qw & \qw                                          & \qw & \qw  & \qw      & \qw                                           & \qw & \qw      &  \qw & \ctrl {1}  & \qw & \qw & \qw & \ctrl{1} \\
& \gate{H} & \qw & \gate{\text{Rz($\boldsymbol{\theta}_{n}$)}}   & \qw & \qw      & \qw & \qw                                          & \qw & \qw  & \qw      & \qw                                           & \qw & \qw      &  \qw & \targ & \qw & \gate{\text{Rz($\boldsymbol{\theta}_{2n-1}$)}} & \qw & \targ
} \]
    \caption{Layer ZZFM.}
    \label{fig:layerZZFM}
\end{figure*}

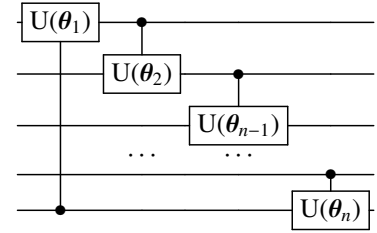
\begin{figure}[H]
    \centering
\[ \Qcircuit @C=0.17em @R=.5em {
&  \gate{\text{U($\boldsymbol{\theta}_{1}$)}} & \ctrl{1} & \qw & \qw      & \qw      & \qw \\
&  \qw                                          &  \gate{\text{U($\boldsymbol{\theta}_{2}$)}}    & \qw & \ctrl{1} & \qw      & \qw \\
&  \qw                                          & \qw      & \qw &  \gate{\text{U($\boldsymbol{\theta}_{n-1}$)}}   & \qw      & \qw \\
&      & \cdots                                         &  &                                                  \cdots \\ 
&  \qw                                          & \qw      & \qw & \qw      & \ctrl{1} & \qw\\
& \ctrl{-5}                                     & \qw & \qw      & \qw &  \gate{\text{U($\boldsymbol{\theta}_{n}$)}}    & \qw
} \]
    \caption{Layer Circular.}
    \label{fig:layerCircular}
\end{figure}

\begin{figure*}
    \centering
\[ \Qcircuit @C=0.17em @R=.4em {
&  \ctrl{1} &  \ctrl{2}   &  \qw &  \ctrl{4}   & \qw & \qw & \ctrl{5} & \qw & \qw & \qw & \qw \\
&  \gate{\text{U($\boldsymbol{\theta}_{1}$)}}     & \qw & \ctrl{1} & \qw & \ctrl{3} & \qw & \qw & \qw & \qw & \ctrl{4} & \qw \\
&  \qw  & \gate{\text{U($\boldsymbol{\theta}_{2}$)}}   & \gate{\text{U($\boldsymbol{\theta}_{3}$)}}     & \qw &  \qw  &  \ctrl{2} & \qw & \ctrl{3} & \qw & \qw & \qw  \\
&      & \cdots                                         &  &                       & &        & & &                      \cdots &    & &  & \\ 
&  \qw          & \qw  & \qw & \gate{\text{U($\boldsymbol{\theta}_{C-6}$)}} & \gate{\text{U($\boldsymbol{\theta}_{C-5}$)}} & \gate{\text{U($\boldsymbol{\theta}_{C-4}$)}} & \qw & \qw & \ctrl{1} & \qw & \qw \\
& \qw & \qw     &  \qw & \qw & \qw & \qw & \gate{\text{U($\boldsymbol{\theta}_{C-3}$)}}    &  \gate{\text{U($\boldsymbol{\theta}_{C-2}$)}} & \gate{\text{U($\boldsymbol{\theta}_{C-1}$)}} & \gate{\text{U($\boldsymbol{\theta}_{C}$)}} & \qw 
} \]
    \caption{Layer Full, where $C = \frac{n!}{2!(n-2)!}$ and $n$ is the amount of qubits of the quantum circuit.}
    \label{fig:layerFull}
\end{figure*}
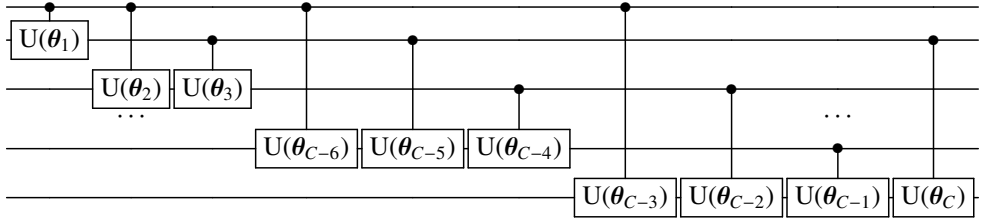

\subsubsection{Quantum variational circuits}

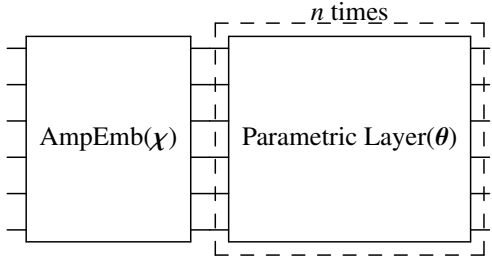
\begin{figure}[H]
    \centering
\[ \Qcircuit @C=0.71em @R=.47em {
&                                 &     &\mbox{$n$ times} \\
&                                 &     & \\
& \multigate{5}{\text{AmpEmb($\boldsymbol{\chi}$)}} & \qw & \multigate{5}{\text{Parametric Layer($\boldsymbol{\theta}$)}} & \qw \\
& \ghost{\text{AmpEmb($\boldsymbol{\chi}$)}}        & \qw & \ghost{\text{Parametric Layer($\boldsymbol{\theta}$)}} & \qw \\
& \ghost{\text{AmpEmb($\boldsymbol{\chi}$)}}        & \qw & \ghost{\text{Parametric Layer($\boldsymbol{\theta}$)}} & \qw \\
& \ghost{\text{AmpEmb($\boldsymbol{\chi}$)}}        & \qw & \ghost{\text{Parametric Layer($\boldsymbol{\theta}$)}} &\qw \\
& \ghost{\text{AmpEmb($\boldsymbol{\chi}$)}}        & \qw & \ghost{\text{Parametric Layer($\boldsymbol{\theta}$)}} &\qw  \\
& \ghost{\text{AmpEmb($\boldsymbol{\chi}$)}}        & \qw & \ghost{\text{Parametric Layer($\boldsymbol{\theta}$)}} &\qw  
 \gategroup{3}{4}{8}{4}{1em}{--}
} \]
    \caption{Generic architecture of a parametric quantum circuit where the Parametric Layer can be repeated $n$ times.}
    \label{fig:genericPQC}
\end{figure}

The variational quantum circuit architecture (or parametric quantum circuit, PQC) was developed in 2014 in \cite{peruzzo2014variational} due to the need to build circuits with few gates, or low depth. A PQC can be divided into two parts: an information loading stage and another parameter loading stage. Figure \ref{fig:genericPQC} presents the representation of a PQC. The parameter loading step can be performed multiple times. Quantum circuits with parameters have been shown to solve complex real-world problems \cite{de2023parametrized, monteiro2021quantum, grant2018hierarchical, de2019implementing}. In \cite{schuld2021effect}, it is discussed that the number of functions that a parametric circuit is capable of modeling increases as repetitions of quantum circuits are performed.
In \cite{ballarin2023entanglement}, results are presented that show that there is convergence of entanglement difference in quantum circuits as there is an increase in the parametric layer. 



There are several ways to load input information \cite{lloyd2020quantum}. One of the most used is loading the amplitude of probabilities of quantum states (called amplitude embedding). Equation \ref{eq:amplitudeembedding} shows the loading of a vector of $N$ values, $\chi_1$, $\chi_2$, ... , $\chi_N$ for a given basis states $\ket{\mathbf{0}}$, $\ket{\mathbf{1}}$, ..., $\ket{\mathbf{N-1}}$.

\begin{equation}
\ket{\psi} = \frac{1}{\sqrt{\sum_b^{N}{\chi_b^2}}} \sum_{b}^{N} \chi_b \ket{\mathbf{b}}  
\label{eq:amplitudeembedding}
\end{equation}

\begin{figure}
    \centering
    \includegraphics[scale=0.5]{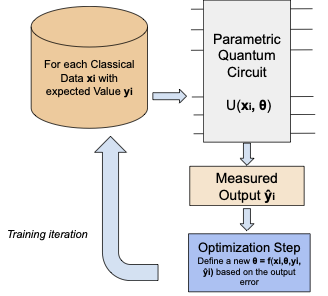}
    \caption{Quantum parametric circuit learning step.}
    \label{fig:learningPQC}
\end{figure}

In terms of parametric circuits, any combination of quantum gates that has rotation parameters is valid.
It is necessary to define a loss function that calculates the circuit error for a given input and a given set of parameters. This loss function is used by the classical optimizer which will adjust the set of parameters to minimize the loss function. Figure \ref{fig:learningPQC} presents a generic diagram of the PQC parameter optimization process.

\section{Experimental protocol}
\label{sec:exp}

\begin{figure*}
    \centering
\includegraphics[scale=0.32]{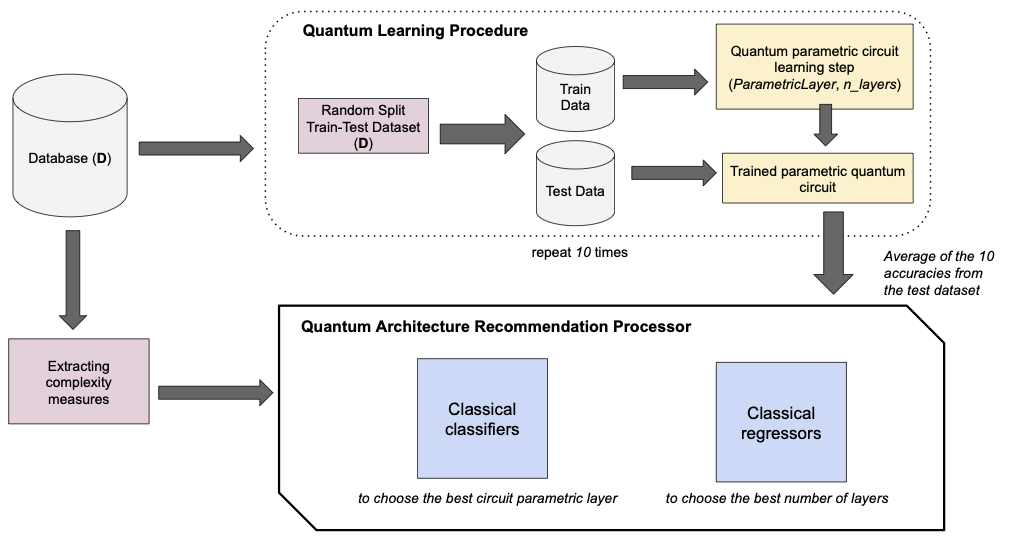}
    \caption{General proposed scheme for recommending quantum circuit architecture.}
    \label{fig:qarecommendation}
\end{figure*}

The experimental protocol overview is seen in Figure \ref{fig:qarecommendation}. 
The quantum architecture recommendation processor (QARP) takes as input the complexity metrics of the databases, as well as the accuracy results and best circuit layer quantities for the 6 possible values of different circuit layers. This results in a table of 14 rows (14 databases) with 22 features (the respective complexity metrics for each database) and a target column (label column) which could be either the best circuit that solved the given problem (largest accuracy on the test set), or the minimum number of layers that a given circuit had to best solve a database.
To find the best results to use in QARP, each of the 14 databases (detailed in Section \ref{sec:datasets}) is run on the 6 classifiers implemented using variational circuits (detailed in Section \ref{sec:parametricscircuits}) with their respective parametrized layer (\textit{parametric\_layer}), which can have 1, 2, 3, 4, 8, or 16 layers (\textit{n\_layers}). The classifiers are trained using the margin loss strategy for data classification, detailed in Section \ref{sec:multiclass}. In each training step, the database is randomly split into training and testing sets in a 70\%/30\% ratio, for a maximum of 100 training epochs. 

To extract complexity metrics from the databases, 10 random splits of the database into training and testing sets (at a 70\%/30\% ratio) are performed, and the average of these complexity measures is taken only from the training sets. 

In this article, one will refer to \textbf{Task~1} as the task of choosing the best parametric circuit for a given task, and \textbf{Task~2} as defining the number of layers in the parametric circuit used. For both tasks, it is possible to use all complexity measures or only one at a time. Classical classifiers and regressors are trained using the complexity metrics of each database as features, with the output (or target) being the best parametric circuit found for that database (\textit{parametric\_layer} for Task 1) and the best number of layers found for that database (\textit{n\_layers} for Task 2). It is important to highlight that in Task 2, the largest number of layers is predicted for the best quantum models. For example, for the Blobs 4F-2C problem, the best circuits found to solve the problem are Layer C-1, Layer C-2, Layer C-3, Layer ZZFM, and Layer Circular (seen in Table \ref{table:bestACCforCircuits}). However, in Table \ref{table:bestQtdeLayerforCircuits}, it is possible to see that the number of layers varies for each of these models. In this way, the largest number of these best models is chosen, which in this example case is 8. This guarantees that the model will predict, in the worst case, more layers, but never less than necessary.

Task 1 can be divided into two tasks depending on how the problem is considered. For example, we can consider that for a problem, only one best circuit is considered to train the QARP. 
Here, we will call it Task 1-A when we consider just one better circuit, which could be Layer C-1, Layer C-2, and so on. But it would also consider all the best circuits by training QARP considering all these circuits. In this case, the task will be called Task 1-B. 


The implementation of the quantum circuit models was carried out using the Pennylane library, in version 0.29, of Python 3.8.8. In the training step, the optimizer used was Adam, from the Pytorch library in version 2.1.2.
The cost function was the Multiclass SVM Loss Function\cite{tang2013deep}, using the parameters margin = 0.15, batch size = 10, and Adam optimizer learning rate = 0.01. Each model was trained 10 times with random splits of the training and testing set. In the end, the average of the maximum accuracy values in the test set was taken.


For training the classical models, the database is divided into training and testing sets. Such random divisions are performed 30 times in each task.

\begin{table*}
\centering
\footnotesize
\caption{Description of the dataset used in the validation experiments of the proposed architecture}
\begin{tabular}{lrrrrrr}
\hline
\textbf{Dataset Name} & \textbf{\# Features} & \textbf{\# Classes} & \textbf{\# Samples} & \textbf{Description}                               \\ 
\midrule
Blobs-1 (B2F2C)       & 2                    & 2                   & 1000                & 2 blobs with cluster standard variation=0.5        \\ 
Blobs-2 (B2F3C        & 2                    & 3                   & 1000                & 3 blobs with cluster standard variation=0.5        \\ 
Blobs-3 (B2F4C)       & 2                    & 4                   & 1000                & 4 blobs with cluster standard variation=0.5        \\ 
Blobs-4 (B4F2C        & 4                    & 2                   & 1000                & 2 blobs with cluster\_std=0.5                      \\ 
Blobs-5 (B4F3C)       & 4                    & 3                   & 1000                & 3 blobs with cluster\_std=0.5                      \\ 
Blobs-6 (B4F4C)       & 4                    & 4                   & 1000                & 4 blobs with cluster\_std=0.5                      \\ 
Circle-1              & 2                    & 2                   & 100                 & 2 concentric circles, scale factor=0.8            \\ 
Circle-2              & 2                    & 2                   & 100                 & 2 concentric circles, scale factor=0.5            \\ 
Moons                 & 2                    & 2                   & 100                 & 2 interleaving half circles                        \\ 
XOR                   & 2                    & 2                   & 2000                & 4 blobs, 2 on opposite sides, being the same class \\ 
Pima Diabetes         & 8                    & 2                   & 768                 &                                                    \\ 
Iris                  & 4                    & 2                   & 150                 &                                                    \\ 
Banknote              & 4                    & 2                   &     1372                &                                                    \\ 
Haberman              & 3                    & 2                   &    306                 & Padding 1 column with 0.1 value                              \\ \bottomrule
\end{tabular}
\label{table:datasetDescription}
\end{table*}

\subsection{Quantum Parametric Circuit Layers}
\label{sec:parametricscircuits}



It will make use of 6 quantum parametric circuit layers, depicted in Figures \ref{fig:layerC1}, \ref{fig:layerC2}, \ref{fig:layerC3}, \ref{fig:layerZZFM}, \ref{fig:layerCircular}, and \ref{fig:layerFull}. These circuits were also analyzed in \cite{ballarin2023entanglement} to assess their levels of entanglement.

For a circuit with $n$ qubits, a single C-1 layer will have $2n$ training parameters. Similarly, a single C-2 layer will have $n$ parameters, as well as the C-3 layer will have $2n-1$, and the ZZFM layer will have $2n-1$. Since generic Rotation U gates have three parameters (as described in Equation \ref{eq:U}), the Circular layer will have $3n$ parameters, as well as the Full layer will have $3 \frac{n!}{2(n-2)!}$ parameters. The number of qubits, $n$, to be used in the circuit depends on the number of features, $\textit{\#features}$, in the database. As the features will be loaded via amplitude embedding, $n=\ceil{log_{2}{\textit{\#features}}}$ qubits will be required. For databases with $n=2$ features, one did not run Layer Circular and Layer Full because they would not explore the complete properties of the arrangement of their layers.

\subsection{Multiclass Margin Quantum classifier}
\label{sec:multiclass}


There are several existing quantum learning algorithms in the literature \cite{li2020quantum}. For this experimental protocol, it was chosen a simple model with low training complexity of implementation. One employs multiple one-vs-all classifiers with a margin loss for data classification. Each classifier is implemented on an individual variational circuit,  consisting of one or multiple layers \cite{tang2013deep, SafwanHossein2020}. This means that one will have a circuit for each of the classes in which the problem is seeking to label inputs.

\subsection{Datasets}
\label{sec:datasets}

\begin{figure*}[b!]
    \centering
  \subfloat[Blobs-B2F2C\label{Blobs-B2F2C}]{%
       \includegraphics[width=0.1782\linewidth]{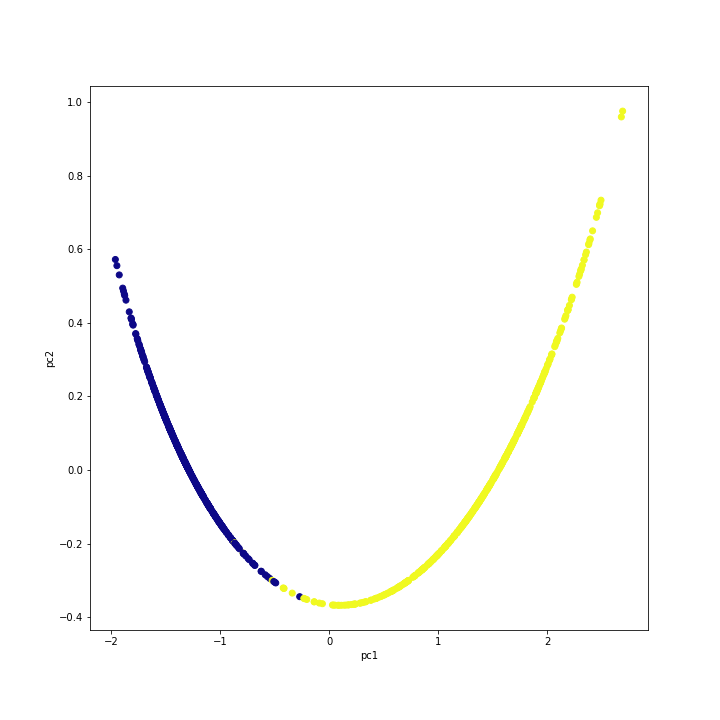}}
  \subfloat[Blobs-B2F3C\label{1b}]{%
        \includegraphics[width=0.1782\linewidth]{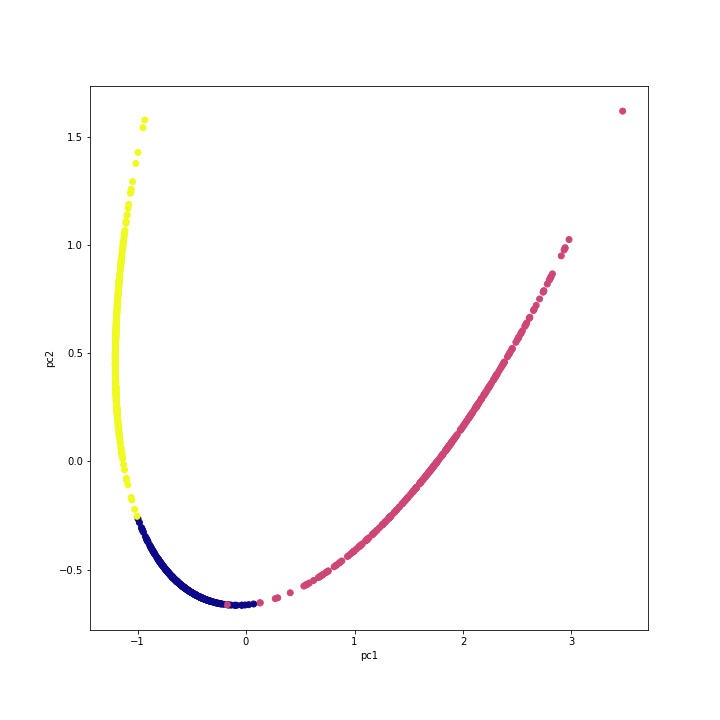}}
  \subfloat[Blobs-B2F4C\label{1c}]{%
        \includegraphics[width=0.1782\linewidth]{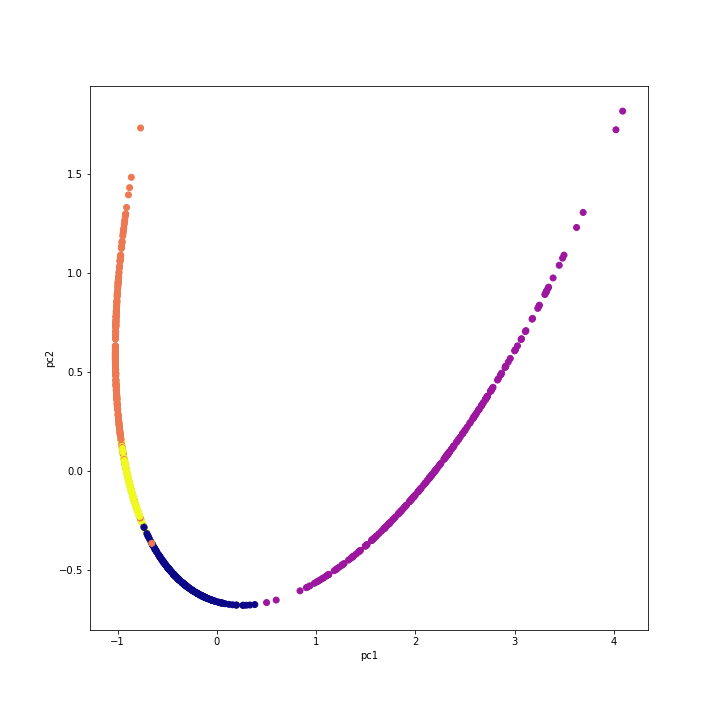}}
  \subfloat[Blobs-B4F2C\label{1d}]{%
        \includegraphics[width=0.1782\linewidth]{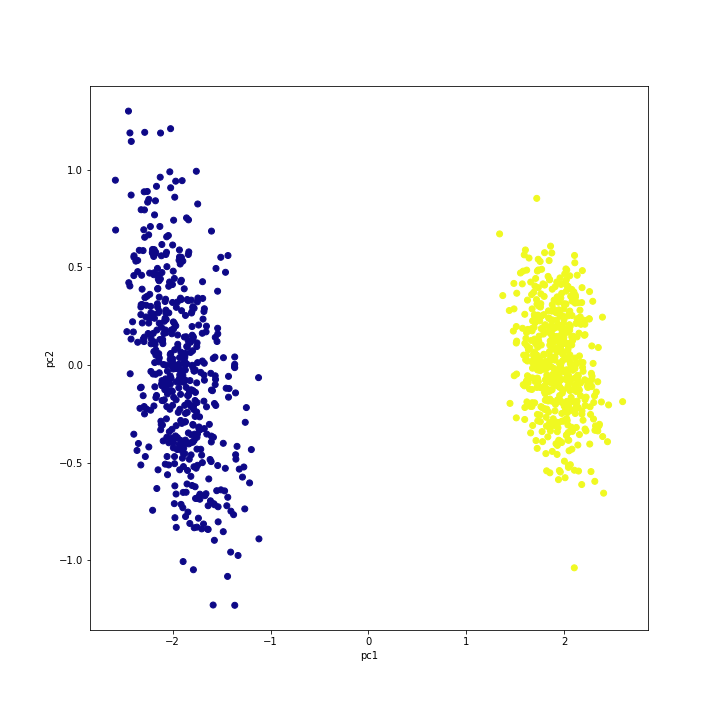}}
    \subfloat[Blobs-B4F3C\label{1d}]{%
        \includegraphics[width=0.1782\linewidth]{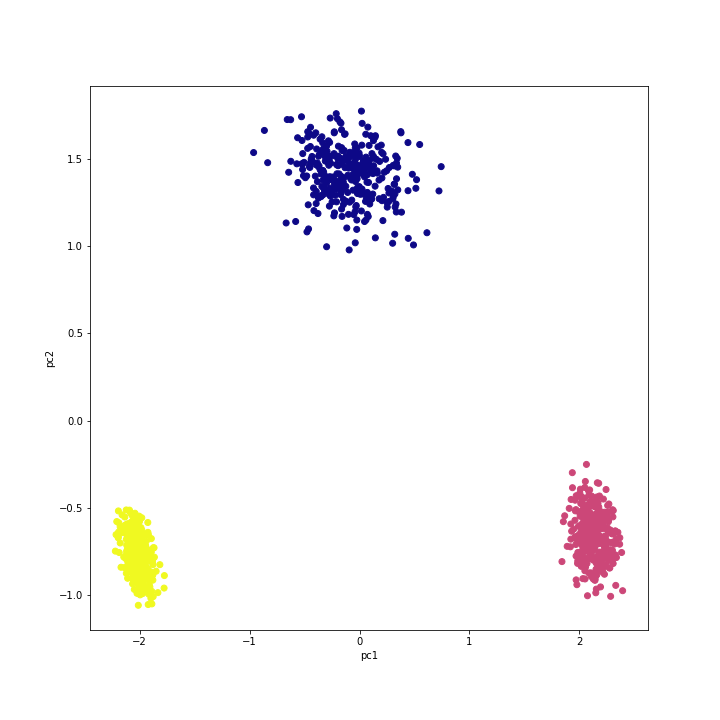}}
    \hfill
    \subfloat[Blobs-B4F4C\label{1d}]{%
        \includegraphics[width=0.1782\linewidth]{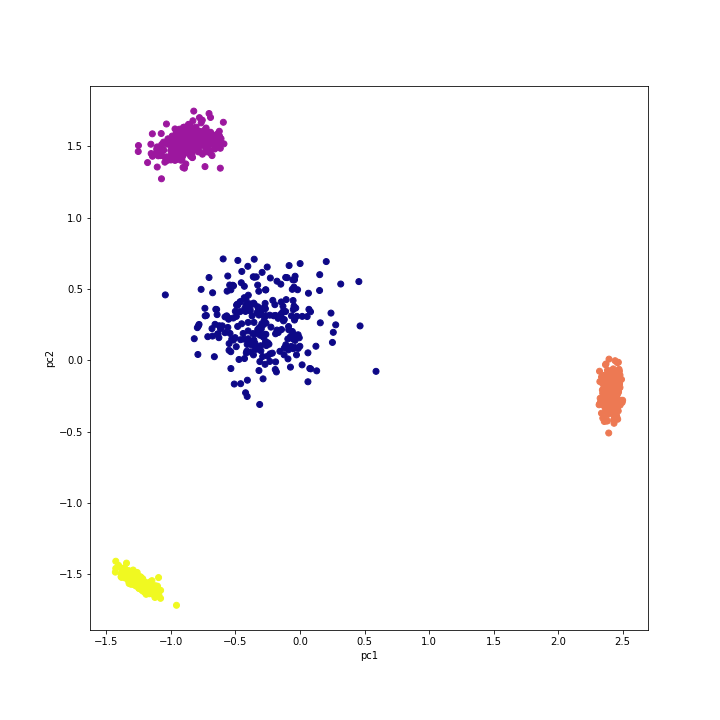}}
    \subfloat[Circle-1\label{1d}]{%
        \includegraphics[width=0.1782\linewidth]{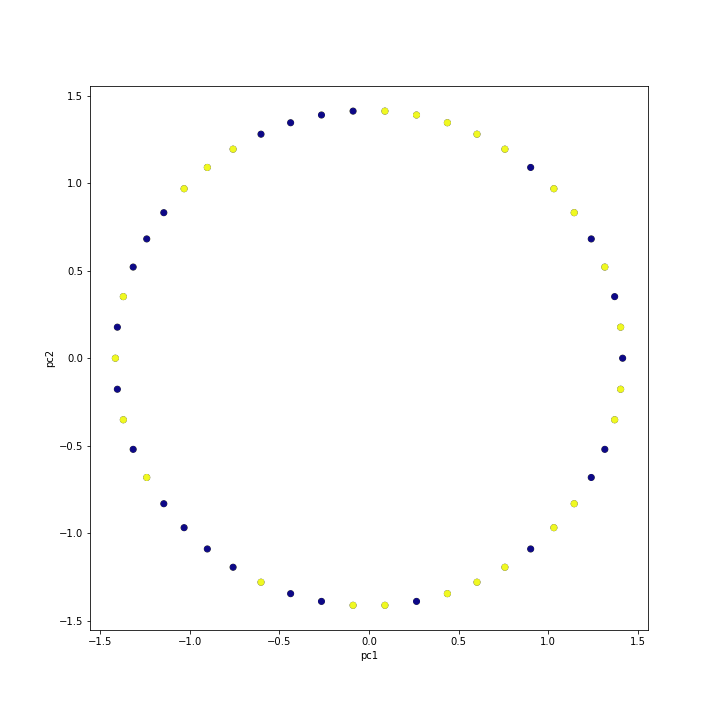}}
    \subfloat[Circle-2\label{1d}]{%
        \includegraphics[width=0.1782\linewidth]{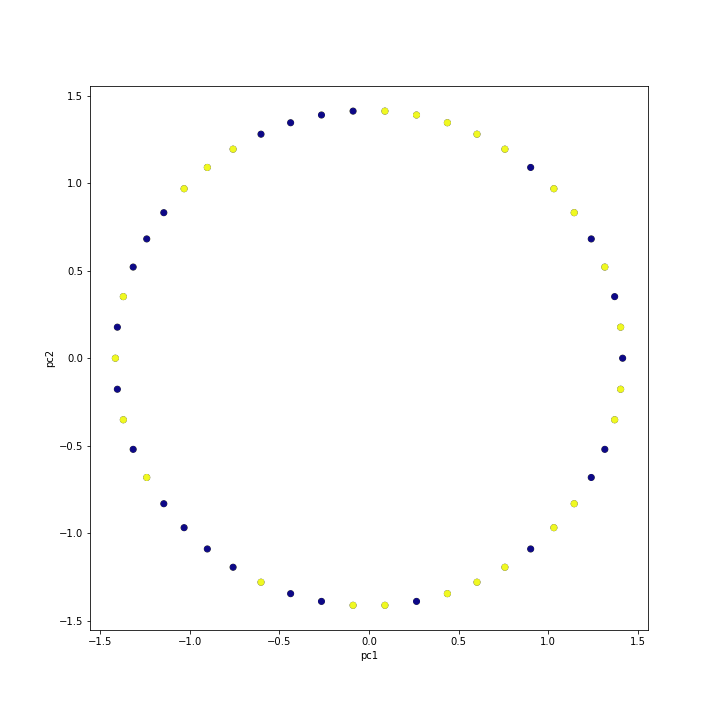}}
    \subfloat[Moons\label{1d}]{%
        \includegraphics[width=0.1782\linewidth]{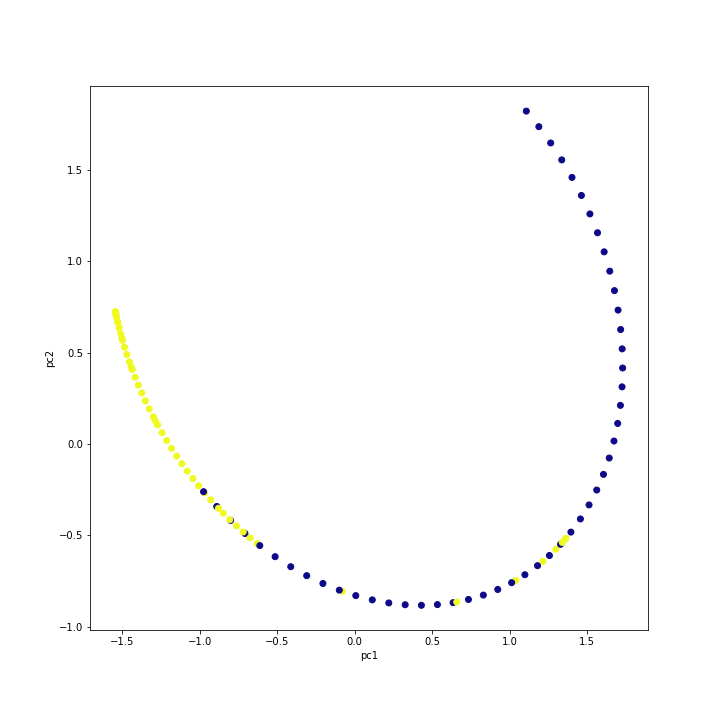}}
    \subfloat[XOR\label{1d}]{%
        \includegraphics[width=0.1782\linewidth]{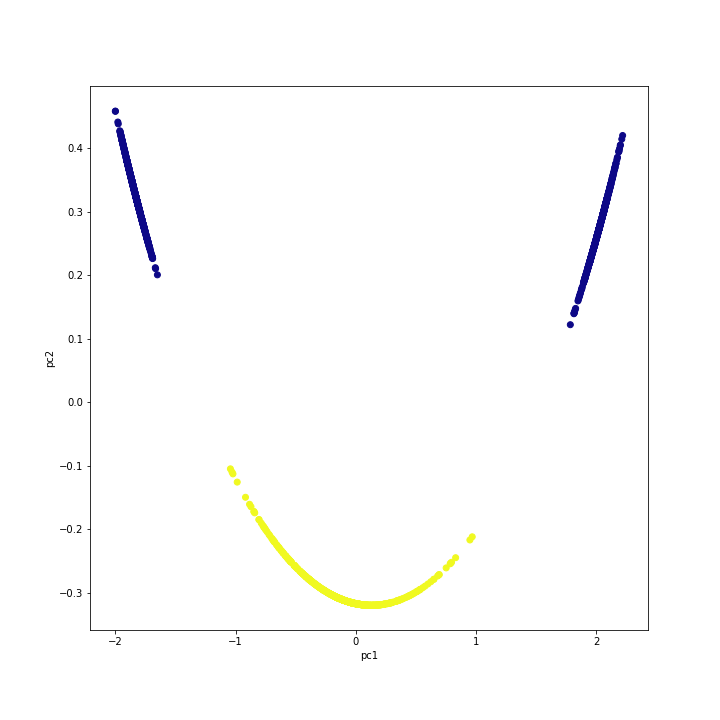}}
    \hfill
    \subfloat[Pima\label{1d}]{%
        \includegraphics[width=0.1782\linewidth]{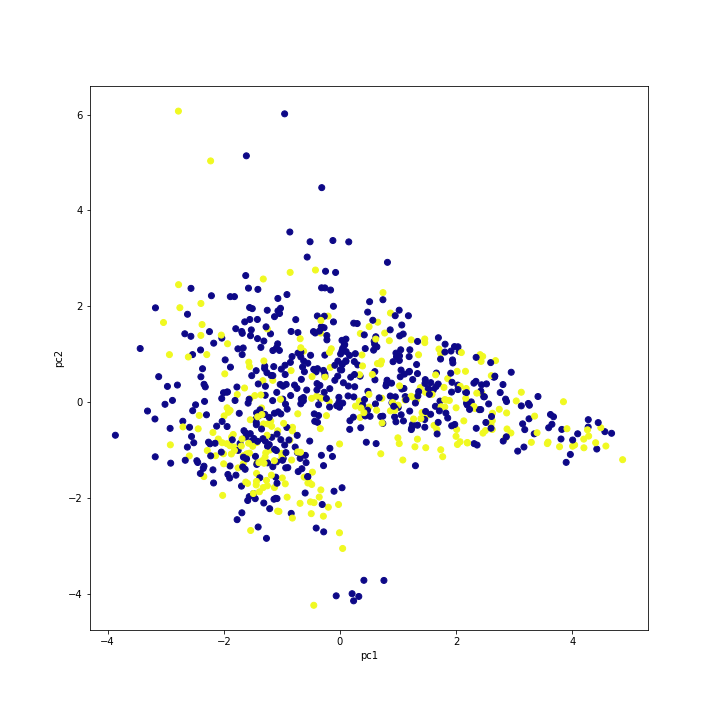}}
    \subfloat[Iris\label{1d}]{%
        \includegraphics[width=0.1782\linewidth]{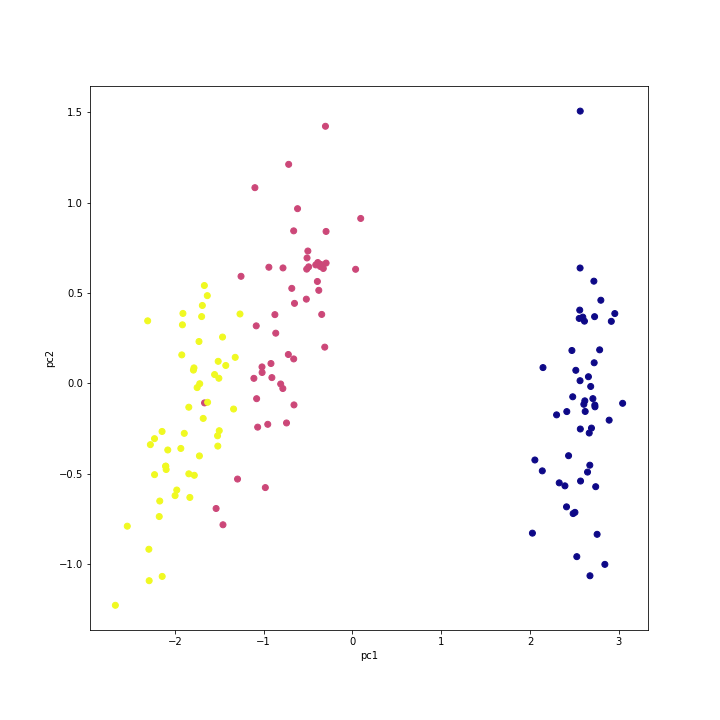}}
    \subfloat[Banknote-ADD\label{1d}]{%
        \includegraphics[width=0.1782\linewidth]{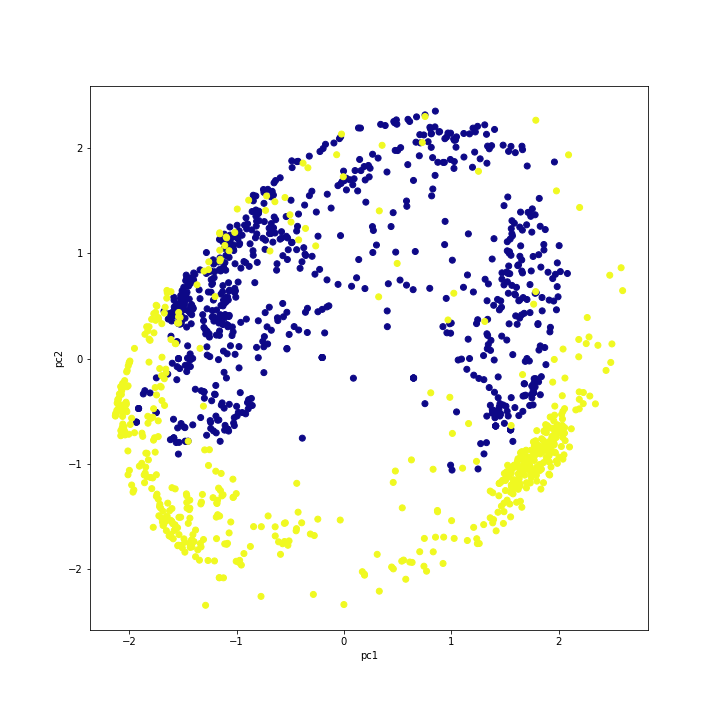}}
    \subfloat[Haberman-ADD\label{1d}]{
        \includegraphics[width=0.1782\linewidth]{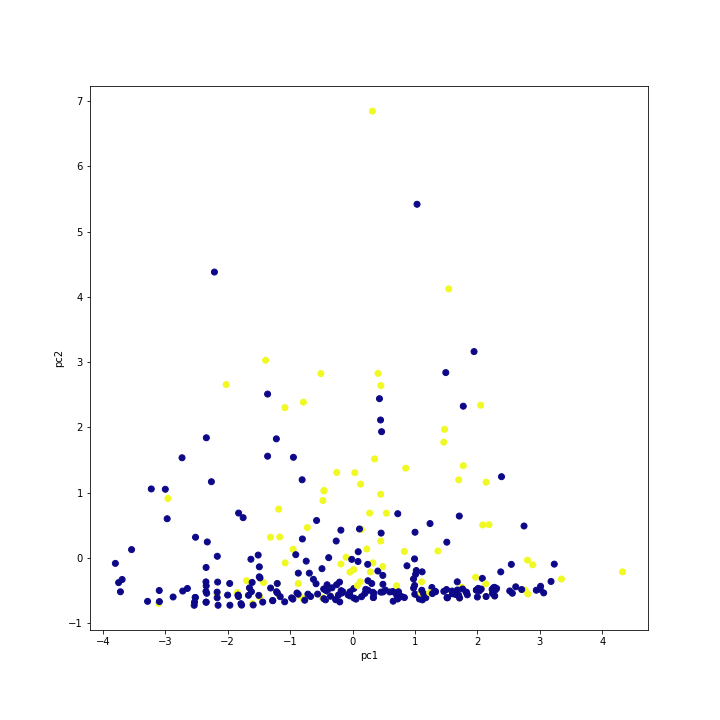}}
  \caption{Visualization of the 12 databases used in the study by PCA decomposition in two dimensions. In each plot, the samples are colored differently depending on their class.}
  \label{fig:basesDeDados} 
\end{figure*}

Fourteen datasets will be executed, with ten of them being synthetic and four real datasets. Their characteristics are detailed in Table \ref{table:datasetDescription}. All features are sample-normalized to a norm of 1, as amplitude encoding requires normalized data. The variability in features and the number of classes in the problems can be observed. The datasets are visualized in Figure~\ref{fig:basesDeDados} in 2 dimensions using Principal Component Analysis (PCA) as a dimensionality reducer. It is worth noting that PCA is solely used for visualization purposes. The quantum circuit experiments utilize all available features from each dataset.

\subsection{Quantum Architecture Recommendation Processor}

The Quantum Architecture Recommendation Processor (QARP) employs classical classifiers and regressors to provide recommendations for the quantum parametric circuit layer and the number of layers in this circuit. Table \ref{table:ModelsForTasks} presents the description and configuration of the models used. The classification and regression models from the Sklearn library version 1.2.2 were utilized \cite{pedregosa2011scikit}. 

QARP receives as input the information contained in Tables \ref{table:bestACCforCircuits} and \ref{table:bestQtdeLayerforCircuits} and 22 database complexity values explained in Section \ref{sec:complexmeasures} and calculated from the Problexity library written in Python available at \cite{komorniczak2023problexity}. The processor recommends a given quantum circuit layout, as well as the layer repetition that improves circuit performance, based on complexity measurements from the database. 

\begin{table*}
\footnotesize
\centering
\caption{Classical Classifiers and Regressors used in the Quantum Architecture Recommendation Processor}
\begin{tabular}{lr}
\hline
Model acronym      & Model name and configuration                                              \\ \hline
DT                 & Decision Tree Classifier                                                             \\
DTR                & Decision Tree Regressor                                                   \\
MLP(R)-500            & Multi-layer Perceptron classifier (or regressor), Hidden layer=500.                       \\
MLP(R)-100-100-100    & Multi-layer Perceptron classifier (or regressor), Hidden layers=100,100,100. \\
MLP(R)-500-500-500    & Multi-layer Perceptron classifier (or regressor), Hidden layers=500,500,500. \\
SVM                & Linear Support Vector Classification.            \\
SVM-RBF            & C-Support Vector Classification, C=1.0, Kernel=RBF, Gamma=Scale.             \\
SVR-RBF            & Epsilon-Support Vector Regression, C=1.0, Kernel=RBF, Gamma=Scale.                 \\
SVM-Sigmoid        & C-Support Vector Classification, C=1.0, Kernel=Sigmoid, Gamma=Scale.         \\
SVR-Sigmoid        & Epsilon-Support Vector Regression, C=1.0, Kernel=Sigmoid, Gamma=Scale.             \\
SVM-Linear         & C-Support Vector Classification, C=1.0, Kernel=Linear, Gamma=scale.          \\
SVR-Linear         & Epsilon-Support Vector Regression, C=1.0, Kernel=Linear, Gamma=Scale.              \\
NaiveBayes         & Gaussian Naive Bayes algorithm for classification.                         \\
kNN                & Classifier implementing the k-nearest neighbors vote, k=5.                 \\
kNNR               & Regression based on k-nearest neighbors, k=5.                              \\
NearestCentroid    & Nearest centroid classifier, metric=euclidean.                            \\
LogisticRegression & Logistic Regression classifier.                                           \\
RF                 & Random forest classifier, n\_estimators=10.                                \\
Ensemble-AB        & AdaBoost classifier, estimator=DecisionTreeClassifier, n\_estimators=50.   \\
Ensemble-Bg        & Bagging classifier, estimator=DecisionTreeClassifier, n\_estimators=10.   \\
Ensemble-GB        & Gradient Boosting for classification.                                     \\
Adaboost           & AdaBoost regressor, estimator = DecisionTreeRegressor, n\_estimators=50.  \\
Bagging            & Bagging classifier. estimator=DecisionTreeClassifier, n\_estimartors=10.  \\
Linear Regression  & Ordinary least squares Linear Regression.                                 \\ \hline
\end{tabular}
\label{table:ModelsForTasks}
\end{table*}

\section{Results}
\label{sec:results}



In the following sections, the results of training classical models for each task will be presented. 

\begin{table*}
\centering
\footnotesize
\caption{Best accuracies for each parameterized layer and each database.}
\begin{tabular}{lrrrrrr}
\toprule
{} &  Layer C-1 &  Layer C-2 &  Layer C-3 &  Layer ZZFM &  Layer Circular &  Layer FULL \\
\midrule
blobs-2F-2C                 &       0.9960       &       0.9960       &       0.9960       &         0.9960        &              -          &            -        \\
blobs-2F-3C                 &       0.9970       &       0.9823       &       0.9823       &         0.9973        &              -          &            -        \\
blobs-2F-4C                 &       0.9617       &       0.8990       &       0.8990       &         0.9597        &              -          &            -        \\
blobs-4F-2C                 &       1.0000       &       1.0000       &       1.0000       &         1.0000        &           1.0000          &         0.5417        \\
blobs-4F-3C                 &       0.9737       &       1.0000       &       1.0000       &         1.0000        &           1.0000          &         0.3497        \\
blobs-4F-4C                 &       0.9763       &       1.0000       &       0.9997       &         1.0000        &           1.0000          &         0.2860        \\
circle-factor-0.5-2F-2C     &       0.6267       &       0.6233       &       0.6233       &         0.5500        &              -          &            -        \\
circle-factor-default-2F-2C &       0.6267       &       0.6233       &       0.6233       &         0.5500        &              -          &            -        \\
moons-2F-2C                 &       0.7500       &       0.7167       &       0.7167       &         0.7433        &              -          &            -        \\
XOR-2F-2C                   &       1.0000       &       1.0000       &       1.0000       &         1.0000        &              -          &            -        \\
PIMA-8F-2C                  &       0.6567       &       0.6810       &       0.6615       &         0.6576        &           0.6532          &         0.6493        \\
Iris                        &       0.9711       &       0.9733       &       0.9756       &         0.9867        &           0.9644          &         0.3933        \\
Haberman                    &       0.7620       &       0.7696       &       0.7609       &         0.7565        &           0.7413          &         0.7348        \\
Banknote                    &       0.7699       &       0.9235       &       0.8459       &         0.9357        &           0.7881          &         0.5779        \\
\bottomrule
\end{tabular}
\label{table:bestACCforCircuits}
\end{table*}

\begin{table*}
    \centering
    \footnotesize
    \caption{Best amount of layer for each parameterized layer and each database.}
    \begin{tabular}{lrrrrrr}
\toprule
{} &  Layer C-1 &  Layer C-2 &  Layer C-3 &  Layer ZZFM &  Layer Circular &  Layer FULL \\
\midrule
blobs-2F-2C                 &              1      &             16      &             16      &               8        &                -         &              -       \\
blobs-2F-3C                 &             16      &             16      &             16      &              16        &                -         &              -       \\
blobs-2F-4C                 &             16      &             16      &             16      &               8        &                -         &              -       \\
blobs-4F-2C                 &              1      &              2      &              8      &               3        &                 2          &               1        \\
blobs-4F-3C                 &              3      &              2      &              8      &               3        &                 2          &               4        \\
blobs-4F-4C                 &              4      &              2      &             16      &               3        &                 8          &               2        \\
circle-factor-0.5-2F-2C     &             16      &             16      &             16      &              16        &                -         &              -       \\
circle-factor-default-2F-2C &             16      &             16      &             16      &              16        &                -         &              -       \\
moons-2F-2C                 &              8      &             16      &             16      &               8        &                -         &              -       \\
XOR-2F-2C                   &              1      &              1      &              1      &               4        &                -         &              -       \\
PIMA-8F-2C                  &              8      &             16      &             16      &              16        &                 8          &               1        \\
Iris                        &              4      &             16      &             16      &               8        &                16          &               4        \\
Haberman                    &             16      &             16      &             16      &              16        &                16          &               8        \\
Banknote                    &             16      &             16      &             16      &              16        &                 8          &               1        \\
\bottomrule
\end{tabular}
    \label{table:bestQtdeLayerforCircuits}
\end{table*}






\subsection{Task 1 - Searching for the Best Quantum Parametric Circuit Layer}

Task 1, the task of choosing the best quantum parametric circuit layer model, can be approached in two simple ways. Either only the best (and simplest) model is considered as the label, or the best (tied) models are considered for the choice.
If the first approach is considered, referred to here as Task 1-A, it will have only 13 examples for training and 1 example for testing. If the second approach is used, referred to here as Task 1-B, it will have 27 examples for training and 2 examples for testing. This is because for each database, sometimes more than one circuit has the best result. 


Table \ref{table:ResultsTask1-1} shows the results of Task 1-A when all database complexity metrics are used to train the classifier models.  
Table \ref{table:ResultsTask1-2} presents the best result found in Task 1-A when only one complexity measure is chosen. In this case, the best measure was N4 using the Ensemble-GB classifier with 0.83 ± 0.37 of average accuracy. 

\begin{table}
\footnotesize
\caption{Results of classical classification models considering average accuracy for solving Task 1-A (searching for the best quantum parametric circuit layer) considering all database complexity metrics.}
\begin{tabular}{lll}
\toprule
{} &    avg $\pm$ std &    min-max \\
\midrule
DT                 &  0.33 $\pm$ 0.47 &  0.00-1.00 \\
SVM                &  0.50 $\pm$ 0.50 &  0.00-1.00 \\
MLP-500            &  \textbf{0.67 $\pm$ 0.47} &  0.00-1.00 \\
MLP-100-100-100    &  0.57 $\pm$ 0.50 &  0.00-1.00 \\
MLP-500-500-500    &  0.50 $\pm$ 0.50 &  0.00-1.00 \\
SVM-RBF            &  0.50 $\pm$ 0.50 &  0.00-1.00 \\
SVM-Sigmoid        &  0.50 $\pm$ 0.50 &  0.00-1.00 \\
SVM-Linear         &  0.50 $\pm$ 0.50 &  0.00-1.00 \\
NaiveBayes         &  0.43 $\pm$ 0.50 &  0.00-1.00 \\
kNN                &  0.50 $\pm$ 0.50 &  0.00-1.00 \\
NearestCentroid    &  0.10 $\pm$ 0.30 &  0.00-1.00 \\
LogisticRegression &  0.40 $\pm$ 0.49 &  0.00-1.00 \\
RF                 &  0.43 $\pm$ 0.50 &  0.00-1.00 \\
Ensemble-AB        &  0.53 $\pm$ 0.50 &  0.00-1.00 \\
Ensemble-Bg        &  0.47 $\pm$ 0.50 &  0.00-1.00 \\
Ensemble-GB        &  0.53 $\pm$ 0.50 &  0.00-1.00 \\
\bottomrule
\end{tabular}
\label{table:ResultsTask1-1}
\end{table}


\begin{table}
\footnotesize
\caption{Results of classical classification models considering average accuracy for solving Task 1-A (search for the best quantum parametric circuit layer) considering the complexity metric N4.}
 \begin{tabular}{lll}
\toprule
{} &    avg $\pm$ std &    min-max \\
\midrule
DT                 &  0.70 $\pm$ 0.46 &  0.00-1.00 \\
SVM                &  0.50 $\pm$ 0.50 &  0.00-1.00 \\
MLP-500            &  0.30 $\pm$ 0.46 &  0.00-1.00 \\
MLP-100-100-100    &  0.43 $\pm$ 0.50 &  0.00-1.00 \\
MLP-500-500-500    &  0.40 $\pm$ 0.49 &  0.00-1.00 \\
SVM-RBF            &  0.50 $\pm$ 0.50 &  0.00-1.00 \\
SVM-Sigmoid        &  0.50 $\pm$ 0.50 &  0.00-1.00 \\
SVM-Linear         &  0.50 $\pm$ 0.50 &  0.00-1.00 \\
NaiveBayes         &  0.53 $\pm$ 0.50 &  0.00-1.00 \\
kNN                &  0.33 $\pm$ 0.47 &  0.00-1.00 \\
NearestCentroid    &  0.37 $\pm$ 0.48 &  0.00-1.00 \\
LogisticRegression &  0.50 $\pm$ 0.50 &  0.00-1.00 \\
RF                 &  0.70 $\pm$ 0.46 &  0.00-1.00 \\
Ensemble-AB        &  0.70 $\pm$ 0.46 &  0.00-1.00 \\
Ensemble-Bg        &  0.67 $\pm$ 0.47 &  0.00-1.00 \\
Ensemble-GB        &  \textbf{0.83 $\pm$ 0.37} &  0.00-1.00 \\
\bottomrule
\end{tabular}
\label{table:ResultsTask1-2}
\end{table}



In Table \ref{table:ResultsTask2_1}, it is shown the results of Task 1-B when considering all the complexity metrics of the databases. Table \ref{table:ResultsTask2_2} displays the best outcome for Task 1-B when only one complexity measure is chosen. In this case, the best measure was T2 with 100\% accuracy. This was the best approach to recommend a quantum circuit for a given problem: we need to consider the best quantum circuits, not just one best, as well as considering only one complexity measure at a time, instead of all measures.

It was possible to find good results using other metrics (and their respective classification models) such as T3 (kNN), F1v (SVM-RBF), L3 (Ensemble-GB), N1 (DT, RF, and Ensemble-GB), N3 (MLP-100-100-100 and SVM-RBF), and N4 (DT and SVM-RBF) with 95\% - 97\% accuracy. T2 is a measure that serves as an indicator of data sparsity. L3, N1, N3, and N4 are indicators of non-linearities and measures of network complexity. It is expected that there will be a correlation of these factors with the necessary complexity of a quantum circuit to solve a given question. The worst accuracy results (and their respective classification models) for this task were found using measures F4 (NearestCentroid) with 0.58 $\pm$ 0.37, C2 (MLP-500-500-500), with 0.63 $\pm$ 0.34, F1 (NearestCentroid) with 0.65 $\pm$ 0.35, LSC (SVM-Sigmoid), C1 (MLP-500) and F3 (Nearest Centroid) with 0.67 $\pm$ 0.32, L1 (NaiveBayes), T1 (NearestCentroid) and CLSCoef (Naive Bayes) with 0.68 $\pm$ 0.35.

\begin{table}
\footnotesize
\caption{Results of classical classification models considering average accuracy for solving Task 1-B (search for the best quantum parametric circuit layer among the best possible) considering all the complexity metrics of the databases.}
\begin{tabular}{lll}
\toprule
{} &    avg $\pm$ std &    min-max \\
\midrule
DT                 &  0.92 $\pm$ 0.19 &  0.50-1.00 \\
SVM                &  0.92 $\pm$ 0.19 &  0.50-1.00 \\
MLP-500            &  \textbf{0.97 $\pm$ 0.12} &  0.50-1.00 \\
MLP-100-100-100    &  0.93 $\pm$ 0.17 &  0.50-1.00 \\
MLP-500-500-500    &  0.95 $\pm$ 0.15 &  0.50-1.00 \\
SVM-RBF            &  0.80 $\pm$ 0.24 &  0.50-1.00 \\
SVM-Sigmoid        &  0.78 $\pm$ 0.25 &  0.50-1.00 \\
SVM-Linear         &  0.90 $\pm$ 0.20 &  0.50-1.00 \\
NaiveBayes         &  0.93 $\pm$ 0.17 &  0.50-1.00 \\
kNN                &  0.85 $\pm$ 0.23 &  0.50-1.00 \\
NearestCentroid    &  0.88 $\pm$ 0.21 &  0.50-1.00 \\
LogisticRegression &  0.92 $\pm$ 0.19 &  0.50-1.00 \\
RF                 &  0.93 $\pm$ 0.17 &  0.50-1.00 \\
Ensemble-AB        &  0.88 $\pm$ 0.21 &  0.50-1.00 \\
Ensemble-Bg        &  0.90 $\pm$ 0.20 &  0.50-1.00 \\
Ensemble-GB        &  0.93 $\pm$ 0.17 &  0.50-1.00 \\
\bottomrule
\end{tabular}
\label{table:ResultsTask2_1} 
\end{table}

\begin{table}
\footnotesize
\caption{Results of classical classification models considering average accuracy for solving Task 1-B (search for the best quantum parametric circuit layer among the best possible) considering the complexity metric T2.}
 \begin{tabular}{lll}
\toprule
{} &    avg $\pm$ std &    min-max \\
\midrule
DT                 &  \textbf{1.00 $\pm$ 0.00} &  1.00-1.00 \\
SVM                &  0.73 $\pm$ 0.28 &  0.00-1.00 \\
MLP-500            &  0.95 $\pm$ 0.15 &  0.50-1.00 \\
MLP-100-100-100    &  0.93 $\pm$ 0.17 &  0.50-1.00 \\
MLP-500-500-500    &  0.92 $\pm$ 0.19 &  0.50-1.00 \\
SVM-RBF            &  0.92 $\pm$ 0.19 &  0.50-1.00 \\
SVM-Sigmoid        &  0.63 $\pm$ 0.29 &  0.00-1.00 \\
SVM-Linear         &  0.77 $\pm$ 0.28 &  0.00-1.00 \\
NaiveBayes         &  0.75 $\pm$ 0.28 &  0.00-1.00 \\
kNN                &  0.88 $\pm$ 0.21 &  0.50-1.00 \\
NearestCentroid    &  0.87 $\pm$ 0.26 &  0.00-1.00 \\
LogisticRegression &  0.73 $\pm$ 0.28 &  0.00-1.00 \\
RF                 &  0.98 $\pm$ 0.09 &  0.50-1.00 \\
Ensemble-AB        &  0.87 $\pm$ 0.22 &  0.50-1.00 \\
Ensemble-Bg        &  0.97 $\pm$ 0.12 &  0.50-1.00 \\
Ensemble-GB        &  0.97 $\pm$ 0.12 &  0.50-1.00 \\
\bottomrule
\end{tabular}
\label{table:ResultsTask2_2}
\end{table}





\subsection{Task 2 - Finding the Best Number of Layers}


For Task 2, as the target value is a numerical one (1,2,3,4,8,16), it is used regressors. Table \ref{table:ResultsTask3_1} shows the results of Task 2 when all complexity metrics of the databases are used as input for the regressor. Table \ref{table:ResultsTask3_2} presents the best result found for Task 2, considering only one complexity measure as input for the regressor. Using all measurements, the best result is 0.67 ± 2.55 mean absolute error, with the DTR regressor; using only one of the measurements at a time, the best result was found using the N2 metric, with the Adaboost algorithm, 0.80 ± 2.17 mean absolute error.
Again, the best approach to indicating the number of circuit layers is one that considers only a measure of complexity, although the error appears with just one more layer. Other metrics (and regressor models) that had low errors were L3 (DTR), with 1.13 $\pm$ 2.42 error, and L2 (Bagging and RF) with 1.23 $\pm$ 2.30, N1 (Adaboost) with 1.33 $\pm$ 2.55, and L1 (DTR) with 1.40 $\pm$ 2.70. L1, L2, and L3 generally measure the distance from expected values considering linear classifiers as a reference. If we deal with the fact that the number of layers has to do with the increasingly non-linear complexity of the model, these measures are completely correlated with the task. Measures N1 and N2 measure the complexity of the database at the data boundary, being an indicator of complexity and measuring the difficulty of the database. 


\begin{table}
\footnotesize
\caption{Results of classical regression models considering the mean absolute error for Task 2 resolution (finding the best number of layers for a quantum circuit) considering all complexity metrics of the databases.}
\begin{tabular}{lll}
\toprule
{} &    MAE $\pm$ std &     min-max \\
\midrule
MLPR-500          &  2.53 $\pm$ 2.19 &   0.00-8.00 \\
MLPR-100-100-100  &  3.23 $\pm$ 2.12 &   1.00-8.00 \\
MLPR-300-300-300  &  2.37 $\pm$ 2.89 &  0.00-11.00 \\
SVR-RBF           &  3.57 $\pm$ 2.60 &   1.00-9.00 \\
SVR-Sigmoid       &  4.47 $\pm$ 2.54 &  0.00-13.00 \\
SVR-Linear        &  3.30 $\pm$ 2.99 &  0.00-10.00 \\
kNNR              &  3.63 $\pm$ 2.04 &   1.00-9.00 \\
Linear Regression &  3.47 $\pm$ 4.13 &  0.00-11.00 \\
DTR               &  \textbf{0.67 $\pm$ 2.55} &  0.00-12.00 \\
Adaboost          &  1.80 $\pm$ 3.52 &  0.00-12.00 \\
Bagging           &  2.67 $\pm$ 2.41 &   0.00-9.00 \\
RF                &  2.30 $\pm$ 2.30 &  0.00-11.00 \\
\bottomrule
\end{tabular}
\label{table:ResultsTask3_1}
\end{table}

\begin{table}
\footnotesize
\caption{Results of classical regression models considering the mean absolute error for Task 2 resolution (finding the best number of layers for a quantum circuit) considering only the complexity metric N2.}
 \begin{tabular}{lll}
\toprule
{} &    MAE $\pm$ std &     min-max \\
\midrule
MLPR-500          &   2.37 $\pm$ 2.73 &   0.00-12.00 \\
MLPR-100-100-100  &   3.97 $\pm$ 1.60 &   2.00-10.00 \\
MLPR-300-300-300  &   2.80 $\pm$ 2.23 &   0.00-10.00 \\
SVR-RBF           &   3.43 $\pm$ 2.32 &    0.00-8.00 \\
SVR-Sigmoid       &  22.33 $\pm$ 6.27 &  12.00-35.00 \\
SVR-Linear        &   3.10 $\pm$ 2.29 &   0.00-10.00 \\
kNNR              &   3.40 $\pm$ 2.04 &    1.00-9.00 \\
Linear Regression &   2.63 $\pm$ 2.02 &   0.00-10.00 \\
DTR               &   1.33 $\pm$ 2.80 &    0.00-8.00 \\
Adaboost          &   \textbf{0.80 $\pm$ 2.17} &    0.00-8.00 \\
Bagging           &   1.63 $\pm$ 2.66 &    0.00-8.00 \\
RF                &   1.60 $\pm$ 2.64 &    0.00-8.00 \\
\bottomrule
\end{tabular}
\label{table:ResultsTask3_2}
\end{table}



The worst mean absolute error results for this Task 2 were found using measures T3 (13.60 $\pm$ 6.22), N2 (22.33 $\pm$ 6.27), CLSCoef (15.33 $\pm$ 8.12), T2 (7.57 $\pm$ 8.59), and F4 (6.20 $\pm$ 5.34).

\section{Conclusion and Future works}
\label{sec:conclusion}

In this work, a quantum circuit recommendation architecture based on the extraction of complexity measures from the database was proposed. The recommendation involves choosing the configuration of the quantum circuit parametric layer (named in the article as Task 1), as well as the number of repetitions that this layer will appear (named in the article as Task 2). The results showed that the architecture can recommend the best circuit out of 6 for the 14 databases used, i.e. with 100\% accuracy, as well as having an error of 0.80 ± 2.17 (i.e., 3 maximum) layers in indicating the number of layers that form the parametric quantum circuit. It was found through experiments that using only one of the database complexity measures for each of these tasks is better than training the classifiers and classical regressors with all measures. The complexity measures and classical models that best helped in these tasks were T2 and Decision Tree (for Task 1) and N2 and Adaboost regressor (for Task 2), respectively.

Future work may provide a chained quantum circuit recommendation with how many layers that quantum circuit should have, using a combination of these measurements to find better results. A proposition of dynamic quantum circuit constructions based on complexity measurements is desired. The evaluation of training hyperameters such as the type of optimizer and its parameter values will be considered in future work. Quantum circuit complexity measurements can also be associated with the calculation of database complexity measurements for the dynamic construction of quantum circuits.


 
%



    \bibliographystyle{elsarticle-num} 
    \bibliography{bib}

\vfill

\end{document}